\documentclass[letterpaper, 10 pt, conference]{ieeeconf}  %

\usepackage{graphicx}\usepackage{booktabs}

\IEEEoverridecommandlockouts                              %

\usepackage{amsmath} %
\usepackage{amssymb} %

\title{\LARGE \bf
InterSocialBench: Benchmarking Human and LLM Preferences for Companion-Robot Social Behavior
}

\author{Yaodan Xu$^{*,\dagger}$, Boyang Guo$^{*}$, Yuqing Gu, Qingxin Zhang,\\
Yiwen Deng, Meng Liu, and Lintian Li$^{\dagger}$%
\thanks{All authors are with Xdream Robotics.}%
\thanks{$^{*}$Yaodan Xu and Boyang Guo contributed equally. $^{\dagger}$Corresponding authors: Yaodan Xu (\texttt{yaodanx@gmail.com}) and Lintian Li (\texttt{lynk@xdreamrobo.com}).}%
}

\usepackage{xcolor}
\usepackage{tikz}
\usetikzlibrary{arrows.meta,positioning}
\definecolor{benchblue}{HTML}{2754A3}
\definecolor{benchteal}{HTML}{287D79}
\definecolor{benchred}{HTML}{B7553C}

\begin{document}
\maketitle
\thispagestyle{empty}
\pagestyle{empty}

\begin{abstract}
Companion robots face everyday situations in which several feasible behaviors may be appropriate, yet different people prefer different responses. We introduce InterSocialBench, a benchmark of 210 domestic scenarios and 18 high-level behaviors, pairing judgments from 100 human participants with 23,520 responses from seven large language models under 16 personality conditions. Each human annotation preserves a preferred action alongside explicitly appropriate and inappropriate candidates. A structured construction pipeline covers behavioral alternatives, competing situational cues, and relevant history and future tasks. Evaluation distinguishes preferred-choice agreement from explicit rejection, using scenario-grouped splits for trainable predictors. Simple frequency and persona-voting baselines illustrate these objectives. Across the tested prompts, model and human behavior distributions differ, and the diversity gap remains after matching response counts: humans exhibit 4.68 distinct choices per scenario, compared with 2.06--3.46 for the models. Human scenario-level plurality agreement is 51.5\%, describing disagreement rather than a universal prediction ceiling. InterSocialBench supports evaluating social behavior selection without replacing individual judgments with a single consensus label.
\end{abstract}

\section{INTRODUCTION}
Companion robots must decide how to respond to people even when no explicit instruction is given. If a user is upset but focused on work, offering comfort and staying quietly nearby may both be reasonable. The appropriate response depends on the situation, and the preferred response can differ across people. A benchmark for this problem must therefore preserve both the alternatives people consider appropriate and the action each person would choose first.

Language models are increasingly used to generate and refine expressive robot behaviors~\cite{mahadevan2024genem,lim2026crisp}, while recent evaluations examine decisions in which human values conflict~\cite{han2026robotvalues}. However, agreement with human social judgments depends on the task and model~\cite{wachowiak2024aligned}. A model may favor behaviors that are common in aggregate while failing to reproduce the diversity of responses within a particular situation. Conversely, failing to predict the first choice of a person does not necessarily mean selecting an explicitly unwanted action. These distinctions motivate our question: how can companion-robot behavior be evaluated against human preferences while preserving reasonable disagreement?

\begin{figure}[!t]
\centering
\includegraphics[width=\columnwidth]{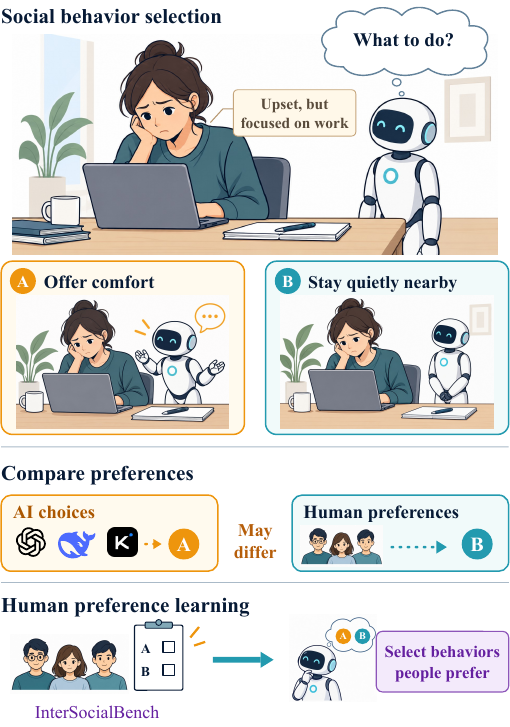}
\caption{Social behavior selection motivated by human preferences. Offering comfort and staying quietly nearby illustrate feasible alternatives whose preference can differ between people and AI. InterSocialBench records human judgments to support evaluation and subsequent preference learning; the depicted choices are illustrative.}
\label{fig:social_behavior_concept}
\end{figure}

InterSocialBench contains 210 domestic scenarios with feasible candidates drawn from 18 behaviors. Scenarios are rendered deterministically from structured specifications describing the current situation, relevant history, and known future tasks. Human participants identify appropriate and inappropriate candidates and choose one highest-priority behavior. We retain these respondent-level labels and compare them with responses from seven LLMs under 16 personality conditions. These conditions provide a controlled collection of model responses, not a random sample of human personalities.

Our contributions are: (i) a benchmark combining structured domestic scenarios with human preferred-choice and appropriateness labels; (ii) a documented construction and annotation protocol covering behavioral alternatives, competing cues, and temporal context; and (iii) simple baseline evaluations and analyses of human--model distribution differences, including response-count-matched diversity. The focus is the benchmark and its evaluation protocol; personalized architectures and physical robot execution are separate research questions.

\section{RELATED WORK}
Social robotics studies systems whose interaction with people is central to their design~\cite{fong2003survey,breazeal2003sociable}. Socially assistive and companion robots extend this interaction to assistance and sustained human--robot relationships~\cite{feilseifer2005defining,dautenhahn2007socially}. Our task concerns the choice among feasible domestic behaviors, rather than the execution of a single instructed task.

Human behavioral evidence already informs robot decisions. MANNERS-DB provides human ratings of robot action appropriateness and neural baselines~\cite{tjomsland2022manners}; GRACE combines LLM predictions with human explanations to generate socially appropriate actions~\cite{dogan2025grace}. Wachowiak et al. compare LLM outputs with human responses across HRI studies, finding task-dependent alignment~\cite{wachowiak2024aligned}. InterSocialBench complements these efforts by combining a preferred action with explicit positive and negative judgments over the same feasible candidates, retaining respondent identity through pseudonymous identifiers, and including structured historical and future context. We do not claim that evaluating social appropriateness itself is new.

Behavior prediction and theory-of-mind approaches address visual behavior modeling, trust-aware interaction, and coordination~\cite{chen2021visual,yu2024toptom,he2025latenttom}. Demonstration and preference learning provide methods for learning robot behavior from human evidence~\cite{argall2009survey,sadigh2017active}, including personalization~\cite{tapus2008personality,gucsi2025useraware} and empirical optimization of interaction~\cite{zhou2022locomotion,slade2024humanloop}. Here, the benchmark supplies observations and individual judgments against which such predictors can be evaluated. It does not require a particular theory-of-mind model or personalization architecture.

\section{TASK AND EVALUATION TARGETS}
\label{sec:problem}
Let $x_q$ describe scenario $q$, including current observations, relevant history, future tasks, and robot resources. The benchmark supplies a feasible candidate set $\mathcal C_q\subseteq\mathcal B$, where $|\mathcal B|=18$. A predictor ranks these candidates and returns a highest-ranked action $\hat b_q$. A valid annotation from respondent $u$ records a preferred action $y_{uq}$, an appropriate set $F_{uq}$, and an inappropriate set $R_{uq}$. Candidates left unmarked are neither explicitly accepted nor explicitly rejected.

For the valid respondent--scenario observations $\mathcal D$, we measure
\begin{equation}
\begin{split}
\mathrm{Top\mbox{-}1}&=\frac{1}{|\mathcal D|}\sum_{(u,q)\in\mathcal D}\mathbf1[\hat b_q=y_{uq}],\\
\mathrm{Reject}&=\frac{1}{|\mathcal D|}\sum_{(u,q)\in\mathcal D}\mathbf1[\hat b_q\in R_{uq}].
\end{split}
\label{eq:metrics}
\end{equation}
Top-3 checks whether $y_{uq}$ occurs in the first three ranked candidates. Accept checks whether $\hat b_q\in F_{uq}$; it is not $1-\mathrm{Reject}$. The Preference--Rejection Profile plots Top-1 against Reject, favoring higher agreement and lower rejection. These are measures of subjective behavioral preference, not physical safety. The benchmark evaluates high-level selection from supplied descriptions and candidates; autonomous perception and motion execution require separate evaluation.

\begin{figure*}[t]
\centering
\includegraphics[width=\textwidth]{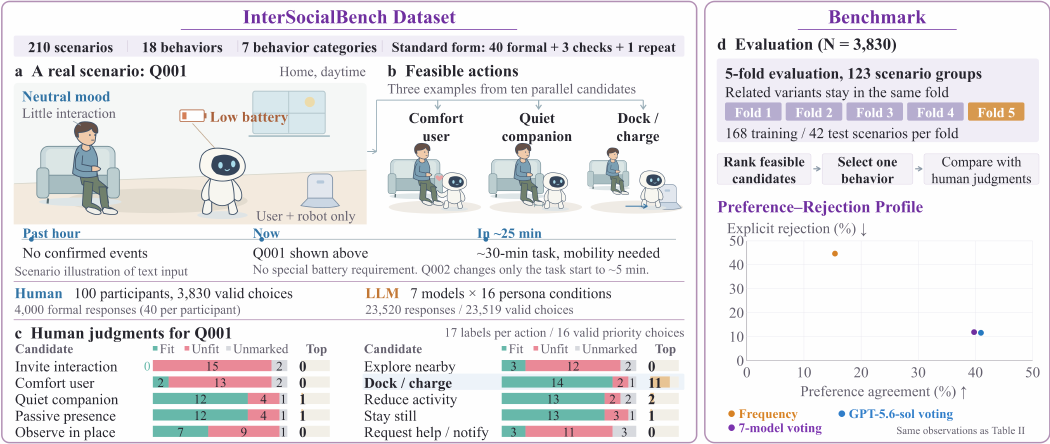}
\caption{InterSocialBench overview. (a) A text scenario illustrated with current and future context. (b) Three of its ten feasible behaviors. (c) Q001 judgments from 17 respondents, including 16 valid preferred choices; fit, unfit, and unmarked counts sum to 17 per action. (d) Scenario-grouped evaluation and three simple baselines from Table~\ref{tab:baselines}: higher preference agreement and lower explicit rejection are better.}
\label{fig:intersocial_benchmark}
\end{figure*}

\section{INTERSOCIALBENCH}
\label{sec:dataset}
InterSocialBench is a questionnaire-based dataset of human judgments and persona-conditioned LLM responses. Figure~\ref{fig:intersocial_benchmark} summarizes its construction and evaluation. The 18 behaviors are grouped by their situational drivers: proactive social contact (invite interaction, maintain interaction, comfort, glance, request/notify); quiet presence (quiet companionship, passive presence); exploration (in-place inspection, local exploration); task continuation; resource management (dock, low activity, idle); self-protection (retreat, defensive hold); and command handling (execute, defer, acknowledge/wait). This mapping is used consistently in construction and category-level analysis; model predictions and primary metrics operate on the 18 specific behaviors.

\subsection{Scenario Construction and Coverage}
\label{sec:construction}
A designer-specified table $A\in\{-2,-1,0,1,2\}^{58\times7}$ maps directional situational cues to behavioral categories. Positive values indicate support, negative values indicate inhibition, and magnitude distinguishes weak and strong effects. These design scores guide which alternatives to include; they are neither human labels nor a learned model of human preferences. For categories $X,Y$, the raw bridge count is
\begin{equation}
B_{\rm raw}(X,Y)=\sum_i\mathbf1[A_{iX}>0\ \land\ A_{iY}>0].
\end{equation}
It counts single cues supporting both categories. The bridge tier $B$ maps counts $0,1,2,3$--$4,5$--$6,\ge7$ to $0,1,2,3,4,5$. A bridge therefore represents multiple plausible alternatives, not necessarily opposing goals.

To measure two-cue tension, let $P_{XY}$ contain distinct, co-occurring cue pairs supporting $X$ and $Y$, respectively; mutually exclusive values of the same input are excluded, and each unordered pair is counted once. Let $O_{XY}\subseteq P_{XY}$ contain pairs where at least one supporting cue also inhibits the other category. We set $T=0$ when $|O_{XY}|/|P_{XY}|<0.20$ or $P_{XY}$ is empty. Otherwise, opposing-pair counts $0,1$--$9,10$--$30,31$--$80,81$--$180,>180$ map to tiers $0,1,2,3,4,5$. For example, negative user emotion supports both social contact and quiet presence. Across all cues, this category pair has four bridge cues ($B=3$) and 132 opposing pairs out of 259 co-occurring pairs ($T=4$). A current invitation and a recent rejection illustrate the two-cue trade-off.

Figure~\ref{fig:conflict_matrix} reports these design tiers. $B+T$ sets a soft allocation priority; it is not a measured magnitude of human disagreement. Within-category allocation uses a separate list of behavior pairs and distinguishing cues: its score is zero for no listed pair and otherwise $\min(5,n+1)$ for $n$ pairs. Thus, the within-category score is not comparable to $B+T$. Density filtering can overlook individual opposing combinations, including low resources with an immediate threat; coverage-driven construction adds such cases rather than interpreting $T=0$ as an absence of possible conflict.

Structured specifications are expanded through event-channel substitutions, recency changes, and background variations, then rendered deterministically into questionnaire text. Checks reject contradictory inputs, require the design-focus action to be feasible, and verify that specified history is actually rendered. Feasibility gates determine candidate availability; designer fit/unfit guesses are not used as human ground truth. Selection prioritizes coverage of situational inputs and all 18 design-focus behaviors, with softer targets for scenario type and cue combinations. The resulting bank contains 63 coverage, 114 cross-category, and 33 within-category scenarios. Cross-category cases include both opposing goals and jointly plausible alternatives.

Table~\ref{tab:quota} distinguishes design targets from realized coverage. Counts of temporal context are obtained from the text actually presented, not construction tags, which can differ after rendering. Variants are not all independent scenarios: the frozen manifest contains 124 source groups, consolidated into 123 evaluation groups to keep linked counterfactuals together. Coverage is designed for diagnostic testing and does not estimate the prevalence of situations in real homes.
\begin{figure}[t]
\centering
\begin{tikzpicture}[x=.61cm,y=.61cm,font=\scriptsize]
\node[anchor=east] at (-.1,0) {Social};\node[rotate=55,anchor=west] at (0,.55) {Social};
\fill[gray!20] (-0.48,-0.48) rectangle (0.48,0.48);\node[text=black] at (0,0) {5};
\fill[benchblue!82] (0.52,-0.48) rectangle (1.48,0.48);\node[text=white] at (1,0) {3/4};
\fill[benchblue!39] (1.52,-0.48) rectangle (2.48,0.48);\node[text=black] at (2,0) {0/3};
\fill[benchblue!39] (2.52,-0.48) rectangle (3.48,0.48);\node[text=black] at (3,0) {0/3};
\fill[benchblue!60] (3.52,-0.48) rectangle (4.48,0.48);\node[text=white] at (4,0) {0/5};
\fill[benchblue!71] (4.52,-0.48) rectangle (5.48,0.48);\node[text=white] at (5,0) {1/5};
\fill[benchblue!71] (5.52,-0.48) rectangle (6.48,0.48);\node[text=white] at (6,0) {3/3};
\node[anchor=east] at (-.1,-1) {Quiet};\node[rotate=55,anchor=west] at (1,.55) {Quiet};
\fill[gray!20] (0.52,-1.48) rectangle (1.48,-0.52);\node[text=black] at (1,-1) {2};
\fill[benchblue!17] (1.52,-1.48) rectangle (2.48,-0.52);\node[text=black] at (2,-1) {1/0};
\fill[benchblue!7] (2.52,-1.48) rectangle (3.48,-0.52);\node[text=black] at (3,-1) {0/0};
\fill[benchblue!49] (3.52,-1.48) rectangle (4.48,-0.52);\node[text=black] at (4,-1) {4/0};
\fill[benchblue!28] (4.52,-1.48) rectangle (5.48,-0.52);\node[text=black] at (5,-1) {2/0};
\fill[benchblue!17] (5.52,-1.48) rectangle (6.48,-0.52);\node[text=black] at (6,-1) {1/0};
\node[anchor=east] at (-.1,-2) {Explore};\node[rotate=55,anchor=west] at (2,.55) {Explore};
\fill[gray!20] (1.52,-2.48) rectangle (2.48,-1.52);\node[text=black] at (2,-2) {2};
\fill[benchblue!17] (2.52,-2.48) rectangle (3.48,-1.52);\node[text=black] at (3,-2) {0/1};
\fill[benchblue!39] (3.52,-2.48) rectangle (4.48,-1.52);\node[text=black] at (4,-2) {1/2};
\fill[benchblue!49] (4.52,-2.48) rectangle (5.48,-1.52);\node[text=black] at (5,-2) {2/2};
\fill[benchblue!7] (5.52,-2.48) rectangle (6.48,-1.52);\node[text=black] at (6,-2) {0/0};
\node[anchor=east] at (-.1,-3) {Task};\node[rotate=55,anchor=west] at (3,.55) {Task};
\fill[gray!20] (2.52,-3.48) rectangle (3.48,-2.52);\node[text=black] at (3,-3) {0};
\fill[benchblue!7] (3.52,-3.48) rectangle (4.48,-2.52);\node[text=black] at (4,-3) {0/0};
\fill[benchblue!7] (4.52,-3.48) rectangle (5.48,-2.52);\node[text=black] at (5,-3) {0/0};
\fill[benchblue!7] (5.52,-3.48) rectangle (6.48,-2.52);\node[text=black] at (6,-3) {0/0};
\node[anchor=east] at (-.1,-4) {Resource};\node[rotate=55,anchor=west] at (4,.55) {Resource};
\fill[gray!20] (3.52,-4.48) rectangle (4.48,-3.52);\node[text=black] at (4,-4) {3};
\fill[benchblue!28] (4.52,-4.48) rectangle (5.48,-3.52);\node[text=black] at (5,-4) {2/0};
\fill[benchblue!28] (5.52,-4.48) rectangle (6.48,-3.52);\node[text=black] at (6,-4) {2/0};
\node[anchor=east] at (-.1,-5) {Protect};\node[rotate=55,anchor=west] at (5,.55) {Protect};
\fill[gray!20] (4.52,-5.48) rectangle (5.48,-4.52);\node[text=black] at (5,-5) {2};
\fill[benchblue!7] (5.52,-5.48) rectangle (6.48,-4.52);\node[text=black] at (6,-5) {0/0};
\node[anchor=east] at (-.1,-6) {Command};\node[rotate=55,anchor=west] at (6,.55) {Command};
\fill[gray!20] (5.52,-6.48) rectangle (6.48,-5.52);\node[text=black] at (6,-6) {4};
\node[anchor=west,align=left,text width=2cm] at (0,-5.1) {Blue: $B/T$ tiers\\Color: $B+T$\\[3pt]Gray: separate\\within-category score};
\end{tikzpicture}
\caption{Design-level bridge/tension tiers ($B/T$). Blue shading encodes allocation priority, not measured human disagreement. Gray diagonal values use the separate within-category rule and are not compared with $B+T$.}
\label{fig:conflict_matrix}
\end{figure}

\begin{table}[t]
\centering
\caption{Realized coverage and nominal design targets.}
\label{tab:quota}
\small\setlength{\tabcolsep}{4pt}
\begin{tabular}{lrrr}
\toprule
Stratum & Count & Share (\%) & Target (\%)\\\midrule
\multicolumn{4}{l}{\textit{Scenario type ($N=210$)}}\\
Coverage & 63 & 30.0 & 30\\
Cross-category & 114 & 54.3 & 50\\
Within-category & 33 & 15.7 & 20\\\midrule
\multicolumn{4}{l}{\textit{Competing cues in cross-category cases ($N=114$)}}\\
Two & 51 & 44.7 & 30\\
Three & 43 & 37.7 & 50\\
Four & 20 & 17.5 & 20\\\midrule
\multicolumn{4}{l}{\textit{Context visible in question text ($N=210$)}}\\
Current only & 103 & 49.0 & --\\
Current + history & 62 & 29.5 & --\\
Current + future & 33 & 15.7 & --\\
Current + both & 12 & 5.7 & --\\\bottomrule
\end{tabular}
\par\smallskip\parbox{\columnwidth}{\footnotesize All 18 behaviors appear as a design focus. Targets are construction preferences, not achieved quotas. Temporal rows describe rendered text and are not compared with tag-based generation targets.}
\end{table}

\subsection{Annotation and Quality Control}
The questionnaire and recorded responses are in Chinese. The standard form contains 40 formal scenarios, two explicit instruction checks, one embedded attention check, and one repeated formal item. The assignment algorithm distributes items across behavioral and temporal contexts. Recency variants and designated counterfactual pairs are assigned to the same respondent and separated in the form; other closely related variants are distributed across respondents.

Annotation targets use disagreement in pilot responses from 112 model--role conditions and three human pilot respondents, distinct from the construction tiers above. Pilot conflict values are mapped by variant identity to the final bank; variants without a match receive the median pilot value. The values are min--max normalized to $c_q\in[0,1]$, and initial targets are $K_q=\mathrm{round}(12+16c_q)$, clipped to $[12,28]$. The assignment script then adjusts targets within these bounds to a total of 4,000 formal responses and constructs forms using seed 20260819. This is a sampling allocation heuristic, not evidence that the construction tiers predict human disagreement.

The standard attention policy requires at least two of three checks to pass; the repeated item is an audit signal, not an exclusion criterion. The final data package retains all 100 submitted participants, including one legacy form without exported checks and one manually reviewed attention-check exception. Attention items and repeated displays are excluded from the formal dataset. This yields 4,000 respondent--scenario observations and 41,003 candidate-level records. Of these observations, 107 lack a preferred choice and 63 mark that choice as inappropriate; the remaining 3,830 form the common evaluation set. Candidate-level judgments from the other observations remain distinguishable in the data. Each scenario has 12--28 formal responses and 11--28 valid preferred choices.

Seven LLMs supply 23,520 scenario responses under 16 persona conditions; one response lacks a unique valid preferred choice, leaving 23,519 for choice-based analysis. Human judgments remain the evaluation reference, and neither designer intentions nor answer-bearing metadata are supplied as model inputs.

For a concrete example, Q001 describes a low-battery robot with a mobility task starting in approximately 25 minutes. Of 17 respondents, 14 mark docking appropriate, two mark it inappropriate, and one leaves it unmarked. There are 16 valid preferred choices, of which 11 select docking. Quiet companionship is marked appropriate by 12 respondents but preferred by only one. Thus, acceptable alternatives and the highest-priority action encode different information. Q002 moves the task start to approximately five minutes while keeping the other stated conditions fixed.

\subsection{Evaluation Protocol}
We use the frozen five-fold manifest (seed 20260817), with 42 scenarios per fold and all related variants in the same one of 123 split groups. A learned or frequency-based predictor may use human labels from the other four folds only. Each valid observation is evaluated once out of fold; the primary scores pool all 3,830 observations. The same respondents can occur across folds, so this split tests transfer to held-out scenario groups, not to unseen users. No individual calibration is used by the simple baselines here.

Distributional summaries instead give each of the 210 scenarios equal weight, preventing scenarios with more respondents from dominating. Plurality agreement compares model choices with the most frequent human action in each scenario; ties use the fixed candidate order. Human plurality fractions describe the concentration of observed choices, not a theoretical ceiling on personalized prediction. The full-data empirical top-one and top-three human actions account for 49.1\% and 83.8\% of valid observations; these descriptive references use the evaluation labels themselves and are not held-out baselines.

\section{EXPERIMENTS AND RESULTS}
\label{sec:experiments}
\subsection{Simple Baselines}
\label{sec:setup}
The seven recorded model sources are GPT-5.6-sol, GLM-5.2, MiniMax-M3, MiMo v2.5-pro, Qwen3.8-Max, Kimi K2.7, and DeepSeek v4-pro. Model names follow the experiment inventory; the response-export identifiers are retained for traceability.

We evaluate four simple baselines on the common observation set. Uniform random ranks feasible candidates uniformly without replacement; scores are exact expectations. Training frequency ranks feasible actions by their preferred-choice counts in the other four folds, ignoring scenario features. GPT-5.6-sol persona voting ranks actions by their collected persona votes, and seven-model voting pools the available votes across all sources. Vote ties use the supplied candidate order, including ties for lower-ranked actions. The voting baselines use no human training labels; their collected scenario responses are not evidence of generalization from training scenarios.

Table~\ref{tab:baselines} demonstrates why preference agreement and rejection are reported jointly. GPT-5.6-sol voting has higher Top-1 agreement than pooled voting, whereas pooled voting has higher Top-3 agreement. Both outperform the context-free frequency reference in preferred-choice matching and explicit rejection. These are descriptive baseline results, not claims about a generally superior model architecture.

\begin{table*}[t]
\centering
\caption{Simple baselines on the same 3,830 valid human observations (\%).}
\label{tab:baselines}
\small\setlength{\tabcolsep}{10pt}
\begin{tabular}{lrrrr}
\toprule
Method & Top-1 $\uparrow$ & Top-3 $\uparrow$ & Accept $\uparrow$ & Reject $\downarrow$\\\midrule
Uniform random (expected) & 10.3 & 31.0 & 48.4 & 37.3\\
Training-fold frequency & 15.4 & 39.1 & 40.2 & 44.7\\
GPT-5.6-sol persona voting & \textbf{41.0} & 68.2 & 81.5 & \textbf{11.7}\\
Seven-model persona voting & 39.8 & \textbf{71.5} & \textbf{82.2} & 12.0\\
\bottomrule
\end{tabular}
\par\smallskip\parbox{.94\textwidth}{\footnotesize Scores pool observations across the frozen five folds. Only the frequency baseline uses human training labels. Voting uses the collected persona responses; ties follow the fixed candidate order. Accept and Reject are separate explicit labels; unmarked predictions contribute to neither. Bold marks the best descriptive value per column, not a significance test.}
\end{table*}

\subsection{Human and Model Response Profiles}
\label{res:r1}
For each source, we calculate its seven-category choice distribution within each scenario and average over scenarios. Total variation is $\mathrm{TV}=\tfrac12\sum_c|p_c-p_c^{\mathrm{human}}|$. Table~\ref{tab:r1} uses the same category mapping as the construction protocol. Under this mapping, quiet presence accounts for 15.4\% of human choices and 17.9--32.6\% of model choices. GPT-5.6-sol has the smallest observed TV (0.086); Qwen3.8-Max has the largest (0.172). These differences characterize the tested scenario bank and prompts, not population-wide traits of model families.

Within-scenario diversity uses the number of distinct specific actions and Shannon entropy $H=-\sum_b p_b\log_2p_b$. In the full samples, humans average 4.96 distinct choices and 1.77 bits per scenario, compared with 2.07--3.50 choices and 0.62--1.33 bits for the models. To separate this difference from response-count effects, for each scenario we set $m_q$ to the smallest valid response count across humans and all seven model sources (11--16). We calculate the exact expected richness and plug-in entropy under uniform subsampling without replacement to $m_q$, then average across scenarios. This analytical rarefaction averages over possible subsets rather than selecting one random subset.

\begin{figure*}[t]
\centering
\begin{tikzpicture}[x=1cm,y=.52cm,font=\footnotesize]
\node[anchor=west,font=\small\bfseries] at (0,1.2) {(a) Distinct actions, matched counts};
\draw[gray!25] (2.5,.5)--(2.5,-7.5);\node[anchor=north] at (2.5,-7.65) {0};
\draw[gray!25] (3.34,.5)--(3.34,-7.5);\node[anchor=north] at (3.34,-7.65) {1};
\draw[gray!25] (4.18,.5)--(4.18,-7.5);\node[anchor=north] at (4.18,-7.65) {2};
\draw[gray!25] (5.02,.5)--(5.02,-7.5);\node[anchor=north] at (5.02,-7.65) {3};
\draw[gray!25] (5.859999999999999,.5)--(5.859999999999999,-7.5);\node[anchor=north] at (5.859999999999999,-7.65) {4};
\draw[gray!25] (6.7,.5)--(6.7,-7.5);\node[anchor=north] at (6.7,-7.65) {5};
\node[anchor=east] at (2.38,0) {Human};
\fill[benchred!75] (2.5,-0.22) rectangle (6.430120671540363,0.22);
\draw[thick] (6.204705125902229,0)--(6.642938273416177,0);
\draw[thick] (6.204705125902229,-0.1)--(6.204705125902229,0.1);\draw[thick] (6.642938273416177,-0.1)--(6.642938273416177,0.1);
\node[anchor=west] at (6.722938273416177,0) {4.68};
\node[anchor=east] at (2.38,-1) {GPT-5.6-sol};
\fill[benchblue!75] (2.5,-1.22) rectangle (4.614821428571428,-0.78);
\draw[thick] (4.425093806509454,-1)--(4.797240967261905,-1);
\draw[thick] (4.425093806509454,-1.1)--(4.425093806509454,-0.9);\draw[thick] (4.797240967261905,-1.1)--(4.797240967261905,-0.9);
\node[anchor=west] at (4.877240967261905,-1) {2.52};
\node[anchor=east] at (2.38,-2) {GLM-5.2};
\fill[benchblue!75] (2.5,-2.22) rectangle (5.140690476190477,-1.78);
\draw[thick] (4.902670422318484,-2)--(5.386506420038229,-2);
\draw[thick] (4.902670422318484,-2.1)--(4.902670422318484,-1.9);\draw[thick] (5.386506420038229,-2.1)--(5.386506420038229,-1.9);
\node[anchor=west] at (5.466506420038229,-2) {3.14};
\node[anchor=east] at (2.38,-3) {MiniMax-M3};
\fill[benchblue!75] (2.5,-3.22) rectangle (4.841833333333334,-2.78);
\draw[thick] (4.6546920646133,-3)--(5.024782370797011,-3);
\draw[thick] (4.6546920646133,-3.1)--(4.6546920646133,-2.9);\draw[thick] (5.024782370797011,-3.1)--(5.024782370797011,-2.9);
\node[anchor=west] at (5.104782370797011,-3) {2.79};
\node[anchor=east] at (2.38,-4) {MiMo v2.5-pro};
\fill[benchblue!75] (2.5,-4.22) rectangle (4.863212087912087,-3.78);
\draw[thick] (4.653458924895228,-4)--(5.058319444233927,-4);
\draw[thick] (4.653458924895228,-4.1)--(4.653458924895228,-3.9);\draw[thick] (5.058319444233927,-4.1)--(5.058319444233927,-3.9);
\node[anchor=west] at (5.138319444233927,-4) {2.81};
\node[anchor=east] at (2.38,-5) {Kimi K2.7};
\fill[benchblue!75] (2.5,-5.22) rectangle (5.405205311355312,-4.78);
\draw[thick] (5.230073086880552,-5)--(5.578138051939003,-5);
\draw[thick] (5.230073086880552,-5.1)--(5.230073086880552,-4.9);\draw[thick] (5.578138051939003,-5.1)--(5.578138051939003,-4.9);
\node[anchor=west] at (5.658138051939003,-5) {3.46};
\node[anchor=east] at (2.38,-6) {DeepSeek v4-pro};
\fill[benchblue!75] (2.5,-6.22) rectangle (4.2312,-5.78);
\draw[thick] (4.0931628685657175,-6)--(4.370357403409091,-6);
\draw[thick] (4.0931628685657175,-6.1)--(4.0931628685657175,-5.9);\draw[thick] (4.370357403409091,-6.1)--(4.370357403409091,-5.9);
\node[anchor=west] at (4.450357403409091,-6) {2.06};
\node[anchor=east] at (2.38,-7) {Qwen3.8-Max};
\fill[benchblue!75] (2.5,-7.22) rectangle (4.742497619047619,-6.78);
\draw[thick] (4.576038823140694,-7)--(4.915397004518278,-7);
\draw[thick] (4.576038823140694,-7.1)--(4.576038823140694,-6.9);\draw[thick] (4.915397004518278,-7.1)--(4.915397004518278,-6.9);
\node[anchor=west] at (4.995397004518278,-7) {2.67};
\node[anchor=west,font=\small\bfseries] at (7.7,1.2) {(b) Entropy (bits), matched counts};
\draw[gray!25] (10.2,.5)--(10.2,-7.5);\node[anchor=north] at (10.2,-7.65) {0};
\draw[gray!25] (11.25,.5)--(11.25,-7.5);\node[anchor=north] at (11.25,-7.65) {0.5};
\draw[gray!25] (12.299999999999999,.5)--(12.299999999999999,-7.5);\node[anchor=north] at (12.299999999999999,-7.65) {1};
\draw[gray!25] (13.35,.5)--(13.35,-7.5);\node[anchor=north] at (13.35,-7.65) {1.5};
\draw[gray!25] (14.399999999999999,.5)--(14.399999999999999,-7.5);\node[anchor=north] at (14.399999999999999,-7.65) {2};
\node[anchor=east] at (10.08,0) {Human};
\fill[benchred!75] (10.2,-0.22) rectangle (13.820117388207626,0.22);
\draw[thick] (13.565646594093456,0)--(14.044834709888063,0);
\draw[thick] (13.565646594093456,-0.1)--(13.565646594093456,0.1);\draw[thick] (14.044834709888063,-0.1)--(14.044834709888063,0.1);
\node[anchor=west] at (14.124834709888063,0) {1.72};
\node[anchor=east] at (10.08,-1) {GPT-5.6-sol};
\fill[benchblue!75] (10.2,-1.22) rectangle (12.109067415497519,-0.78);
\draw[thick] (11.846760040655994,-1)--(12.366456769297459,-1);
\draw[thick] (11.846760040655994,-1.1)--(11.846760040655994,-0.9);\draw[thick] (12.366456769297459,-1.1)--(12.366456769297459,-0.9);
\node[anchor=west] at (12.446456769297459,-1) {0.91};
\node[anchor=east] at (10.08,-2) {GLM-5.2};
\fill[benchblue!75] (10.2,-2.22) rectangle (12.601565584634688,-1.78);
\draw[thick] (12.294611491119362,-2)--(12.899868703196416,-2);
\draw[thick] (12.294611491119362,-2.1)--(12.294611491119362,-1.9);\draw[thick] (12.899868703196416,-2.1)--(12.899868703196416,-1.9);
\node[anchor=west] at (12.979868703196416,-2) {1.14};
\node[anchor=east] at (10.08,-3) {MiniMax-M3};
\fill[benchblue!75] (10.2,-3.22) rectangle (12.25340031122367,-2.78);
\draw[thick] (12.017578935825517,-3)--(12.481327701340748,-3);
\draw[thick] (12.017578935825517,-3.1)--(12.017578935825517,-2.9);\draw[thick] (12.481327701340748,-3.1)--(12.481327701340748,-2.9);
\node[anchor=west] at (12.561327701340748,-3) {0.98};
\node[anchor=east] at (10.08,-4) {MiMo v2.5-pro};
\fill[benchblue!75] (10.2,-4.22) rectangle (12.368645660418347,-3.78);
\draw[thick] (12.08498700236617,-4)--(12.62386026807053,-4);
\draw[thick] (12.08498700236617,-4.1)--(12.08498700236617,-3.9);\draw[thick] (12.62386026807053,-4.1)--(12.62386026807053,-3.9);
\node[anchor=west] at (12.70386026807053,-4) {1.03};
\node[anchor=east] at (10.08,-5) {Kimi K2.7};
\fill[benchblue!75] (10.2,-5.22) rectangle (12.976655346329567,-4.78);
\draw[thick] (12.741750159789362,-5)--(13.194120519841057,-5);
\draw[thick] (12.741750159789362,-5.1)--(12.741750159789362,-4.9);\draw[thick] (13.194120519841057,-5.1)--(13.194120519841057,-4.9);
\node[anchor=west] at (13.274120519841057,-5) {1.32};
\node[anchor=east] at (10.08,-6) {DeepSeek v4-pro};
\fill[benchblue!75] (10.2,-6.22) rectangle (11.498383335962524,-5.78);
\draw[thick] (11.289137170970678,-6)--(11.696364428474602,-6);
\draw[thick] (11.289137170970678,-6.1)--(11.289137170970678,-5.9);\draw[thick] (11.696364428474602,-6.1)--(11.696364428474602,-5.9);
\node[anchor=west] at (11.776364428474603,-6) {0.62};
\node[anchor=east] at (10.08,-7) {Qwen3.8-Max};
\fill[benchblue!75] (10.2,-7.22) rectangle (12.171896675283808,-6.78);
\draw[thick] (11.963275681047822,-7)--(12.394290199387072,-7);
\draw[thick] (11.963275681047822,-7.1)--(11.963275681047822,-6.9);\draw[thick] (12.394290199387072,-7.1)--(12.394290199387072,-6.9);
\node[anchor=west] at (12.474290199387072,-7) {0.94};
\end{tikzpicture}
\caption{The human--model diversity gap remains after matching response counts within each scenario. Bars show exact subsampling expectations; whiskers show 95\% scenario-group bootstrap intervals. Both panels use specific actions, not the seven aggregate categories.}
\label{fig:r1}
\end{figure*}
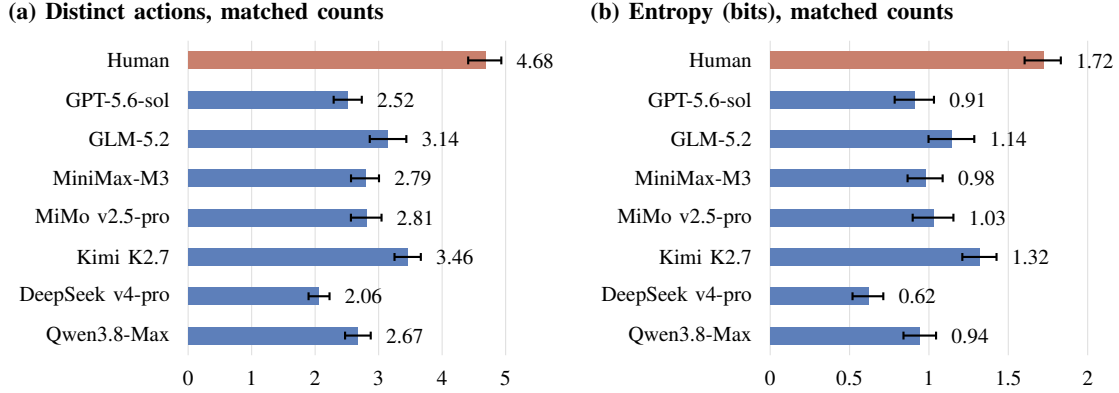

Figure~\ref{fig:r1} shows the count-matched results. Human richness is 4.68 versus 2.06--3.46 for the models; entropy is 1.72 versus 0.62--1.32 bits. Paired bootstrap intervals over the 123 scenario groups (2,000 replicates, seed 20260914) place the mean difference of each model from humans below zero for both measures. These intervals describe variation across scenario groups conditional on the collected respondents and prompts; they do not represent uncertainty from recruiting new participants. The result supports a diversity gap under the tested persona conditions even after matching counts.

Human scenario-averaged plurality fractions are 51.5\% for specific actions and 63.9\% for the seven construction categories. These differ from observation-weighted summaries because each scenario receives one vote. Neither plurality agreement nor category-level similarity establishes that a model reproduces individual preferences.

\begin{table*}[t]
\centering
\caption{Scenario-averaged category profiles (\%) and distance from humans.}
\label{tab:r1}
\small\setlength{\tabcolsep}{5pt}
\begin{tabular}{lrrrrrrrrr}
\toprule
Source & Social & Quiet & Explore & Task & Resource & Protect & Command & TV & Plurality\\\midrule
Human & 29.4 & 15.4 & 5.7 & 3.3 & 27.8 & 12.1 & 6.4 & -- & --\\
\midrule
GPT-5.6-sol & 31.3 & 18.8 & 5.0 & 3.5 & 23.0 & 15.1 & 3.2 & 0.086 & 55.5\\
GLM-5.2 & 22.9 & 18.0 & 5.9 & 3.0 & 33.9 & 12.5 & 3.7 & 0.094 & 50.4\\
MiniMax-M3 & 21.5 & 22.1 & 2.8 & 2.9 & 34.5 & 10.9 & 5.2 & 0.134 & 54.1\\
MiMo v2.5-pro & 18.8 & 17.9 & 5.7 & 3.8 & 37.0 & 13.4 & 3.3 & 0.136 & 51.9\\
Kimi K2.7 & 19.1 & 18.8 & 8.8 & 2.7 & 36.1 & 11.3 & 3.2 & 0.148 & 47.1\\
DeepSeek v4-pro & 18.7 & 28.3 & 2.9 & 2.8 & 30.6 & 12.3 & 4.3 & 0.159 & 58.2\\
Qwen3.8-Max & 18.8 & 32.6 & 4.3 & 3.3 & 24.6 & 10.8 & 5.6 & 0.172 & 49.8\\
\bottomrule
\end{tabular}
\par\smallskip\parbox{.94\textwidth}{\footnotesize Each scenario receives equal weight. Social includes glance and request/notify; resource includes idle. TV uses unrounded proportions. Plurality is agreement (\%) of persona responses with the most frequent human action in the scenario, with ties resolved by fixed candidate order; it is not individual-choice accuracy.}
\end{table*}

\section{DISCUSSION AND LIMITATIONS}
InterSocialBench separates three questions: which action a person prefers, whether a proposed action is explicitly unwanted, and whether a collection of model responses reproduces the distribution of human choices. The baseline table addresses the first two; the response-profile and count-matched diversity analyses address the third. Good performance on one does not establish good performance on all three.

The benchmark is a finite, deliberately constructed bank of text scenarios rather than a representative sample of domestic robot encounters. Its design table, feasibility rules, and coverage priorities encode designer assumptions. Human labels provide an independent response to these scenarios but do not remove those construction choices. The persona-conditioned samples are also not independent draws from a human population; comparisons are conditional on the tested prompts and model versions. A same-prompt repeated-sampling control would be needed to isolate effects of persona conditioning. Scenario-group bootstraps account for related question variants but not uncertainty from sampling a new cohort.

The final cohort includes two documented attention-policy exceptions. Repeated-item consistency is retained as a quality diagnostic rather than a prediction ceiling; the preparation audit records 43 matching and 56 nonmatching repeated top choices, with one unavailable record. A single repeated item per participant cannot establish stable individual preferences or separate inattention from contextual or preference variability. Participants gave informed consent, including permission to share deidentified questionnaire responses. Modeling exports use pseudonymous identifiers with direct identifying fields removed.

\section{CONCLUSION}
InterSocialBench pairs structured companion-robot scenarios with individual human preferences and explicit appropriateness judgments. Simple baselines demonstrate joint preferred-choice and rejection evaluation, while count-matched analyses show less within-scenario choice diversity in the tested LLM responses than in human responses. The benchmark supports reproducible evaluation of behavior selection without treating human disagreement as annotation error or replacing individual preferences with a single majority answer.

\end{document}